\documentclass{article}

\usepackage{microtype}
\usepackage{graphicx}
\usepackage{booktabs} 
\usepackage{xcolor}

\usepackage{hyperref}
\usepackage{url}
\usepackage{graphics}
\usepackage{caption}
\usepackage{babel}
\usepackage{booktabs}
\usepackage{multirow}
\usepackage{wrapfig}
\usepackage[T1]{fontenc}
\usepackage{amsmath}
\usepackage{amssymb}
\usepackage{subcaption}
\usepackage{makecell}

\usepackage{hyperref}

\newcommand{\yonghao}[1]{#1}

\newcommand{\sys}{AlpaComm }

\usepackage[accepted]{mlsys2023}

\mlsystitlerunning{On Optimizing the Communication of Model Parallelism}

\usepackage[firstpage]{draftwatermark}
\SetWatermarkText{
 \hspace*{4.5in}
 \raisebox{10in}{
  \includegraphics[height=0.9in]{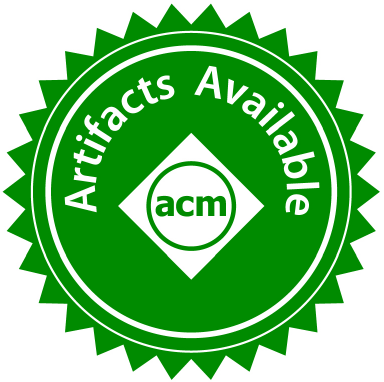}%
  \includegraphics[height=0.9in]{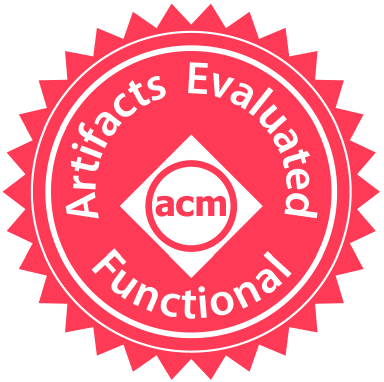}
 }
}
\SetWatermarkAngle{0}

\begin{document}

\twocolumn[
\mlsystitle{On Optimizing the Communication of Model Parallelism}



\mlsyssetsymbol{equal}{*}

\begin{mlsysauthorlist}
\mlsysauthor{Yonghao Zhuang}{equal,cmu}
\mlsysauthor{Hexu Zhao}{equal,tsinghua} 
\mlsysauthor{Lianmin Zheng}{berkeley}
\mlsysauthor{Zhuohan Li}{berkeley}

\mlsysauthor{Eric P. Xing}{cmu,mbzuai,petuum} 
\mlsysauthor{Qirong Ho}{mbzuai,petuum} 
\mlsysauthor{Joseph E. Gonzalez}{berkeley}
\mlsysauthor{Ion Stoica}{berkeley}
\mlsysauthor{Hao Zhang}{berkeley}
\end{mlsysauthorlist}

\mlsysaffiliation{cmu}{Carnegie Mellon University}
\mlsysaffiliation{tsinghua}{Tsinghua University}
\mlsysaffiliation{berkeley}{University of California, Berkeley}
\mlsysaffiliation{mbzuai}{MBZUAI}
\mlsysaffiliation{petuum}{Petuum Inc.}

\mlsyscorrespondingauthor{Yonghao Zhuang}{yhzhuang@cmu.edu}

\mlsyskeywords{Machine Learning, MLSys}

\vskip 0.3in

\begin{abstract}
We study a novel and important communication pattern in model-parallel deep learning (DL) at scale, which we call cross-mesh resharding. This pattern emerges when the two paradigms of model parallelism -- intra-operator and inter-operator parallelism -- are combined to support large models on large clusters.
In cross-mesh resharding, a sharded tensor needs to be sent from a source device mesh to a destination device mesh, on which the tensor may be distributed with the same or different layouts.
We formalize this as a many-to-many multicast communication problem, and show that existing approaches either are sub-optimal or do not generalize to different network topologies or tensor layouts, which result from different model architectures and parallelism strategies. We then propose two contributions to address cross-mesh resharding: an efficient broadcast-based communication system, and an ``overlapping-friendly" pipeline schedule. On microbenchmarks, our overall system outperforms existing ones by up to 10x across various tensor and mesh layouts. 
On end-to-end multi-node training of two models, GPT and U-Transformer, we improve throughput by 10\% and 50\%, respectively.

\end{abstract}
]



\printAffiliationsAndNotice{*Yonghao and Hexu contributed equally. Part of this work was done when Hexu was intern at MBZUAI}  

\section{Introduction}
\begin{figure*}[t]
	\centering
	\includegraphics[width=0.95\textwidth]{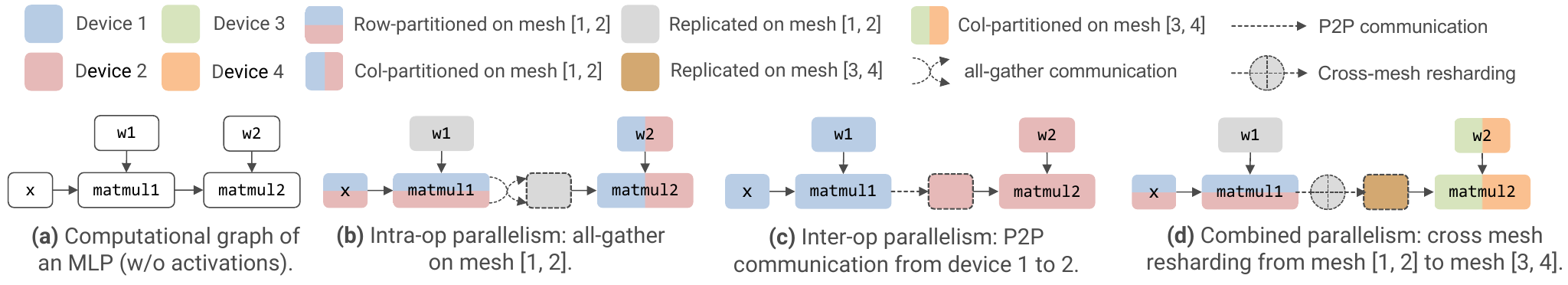}
	\vskip -0.5em
	\caption{Model parallelism applies to an MLP, where each node represents an operator and its output tensor, and solid arrows indicate data flowing directions. The node with dotted boundaries in each of (b)-(d) shows the required input layout to \texttt{matmul2}. When combining two parallelisms in (d), cross-mesh resharding emerges for both exchanging tensors and converting their layouts between two meshes.}
	\vskip -1em
	\label{fig:mp-comm}
\end{figure*}
Model-parallel distributed training and inference using GPU or TPU clusters have been key drivers for many recent advances~\cite{brown2020language,shoeybi2019megatron} in deep learning (DL). Concurrent model parallelism approaches can be roughly classified into two paradigms: intra-op parallelism that partitions individual layers or computational operators of the model~\cite{xu2021gspmd,lepikhin2020gshard} and inter-op parallelism which partitions the computational graphs~\cite{huang2019gpipe,narayanan2021efficient}. 


Core to the performance of these model parallelism approaches is the need to communicate partial results between parallel devices, which are responsible for different parts of the model computations.
Because there are many possible parallelism strategies, there is a correspondingly wide variety of communication patterns that results from said strategies. 
For example, in intra-op parallelism, communication is required to transform a tensor~\cite{xu2021gspmd,wang2019supporting} (which is sharded over a group of devices called a \emph{device mesh}) between two distributed tensor layouts (Figure~\ref{fig:mp-comm}b); whereas, in inter-op parallelism, communication is required to exchange a full tensor between a pair of devices (Figure~\ref{fig:mp-comm}c). 
Both intra-op and inter-op communication can be easily implemented using, respectively, existing collective and point-to-point (P2P) communication primitives, as previous research~\cite{zheng2022alpa,xu2021gspmd} has revealed.

In practice however, neither intra-op nor inter-op parallelism alone suffices to scale out to models like GPT-3~\cite{brown2020language, narayanan2021efficient}. Instead, they must be combined and lead to a new and composite communication pattern, called \emph{cross-mesh resharding}, which requires (1) exchanging a tensor between two different device meshes, and simultaneously (2) changing its layout (see Figure~\ref{fig:mp-comm}d and \S\ref{sec:background:cmr}). Since the communication happens from one group of devices to the other group, it cannot be implemented by directly using existing collective communication primitives. At the same time, because the tensor may have a different layout on the destination mesh, the communication also cannot be directly addressed by P2P communication primitives.
Several previous works~\cite{narayanan2021efficient,zheng2022alpa} have proposed specialized solutions to this problem, but only for transformer architectures under a fixed parallelism plan. As we will show later, none of them are optimal or general -- for example, when they are applied to emerging models such as U-Transformer~\cite{petit2021unettransformer} -- the backbone of diffusion models~\cite{rombach2022high} -- they cause substantial slowdowns in communication.


In this paper, we ask: given an arbitrary model and its composite inter-op and intra-op parallelism plan, what is the best way to perform cross-mesh resharding? We formalize this as a many-to-many multicast communication problem between two device meshes, and reveal several unique challenges in the context of model-parallel DL:



\noindent \textbf{Complex communication requirements.}
The message transferred in cross-mesh resharding is a distributed tensor sharded on two separate meshes. They may exhibit distinct layouts on the source and destination mesh (which consist of different, non-overlapping, sets of devices). For example, a slice of the tensor might be {\it replicated} over the source devices, but devices in the destination mesh may be expecting different parts of that same slice -- in other words, the destination mesh requires the slice to be {\it partitioned}. Fulfilling such communication requirements while minimizing communication costs requires careful choice of communication primitives, sender and receiver devices, and communication routes.


\noindent \textbf{Heterogeneous networking.}
A deep learning cluster contains heterogeneous networking hardware: NVLink, PICe, NIC, InfiniBand, etc.; this results in uneven bandwidth when devices communicate within and across meshes. Thus, in a model-parallel job, we need to carefully consider how to assign communication tasks, in order to make the best use of uneven bandwidth whilst balancing network traffic.




\noindent \textbf{Dependency with computation schedules.}
An entire model-parallel job contains many cross-mesh resharding tasks, which are triggered whenever one mesh's computation depends on another mesh's results. Such computational dependencies are fully defined by the model computation and parallelism plan. To improve the model's overall performance, we need to optimally schedule and overlap these computation and communication tasks.

To address these challenges, we propose new solutions to (1) optimize a single arbitrary cross-mesh resharding task, and (2) collectively optimize all cross-mesh resharding tasks in a model-parallel job. Our contributions are summarized as:

\noindent \(\bigstar\)  We formalize the cross-mesh resharding problem in model-parallel DL jobs, and reveal its characteristics.

\noindent \(\bigstar\) We propose a general and principled solution to optimize for a single cross-mesh resharding task. 
Specifically, we first show that communication primitives in existing solutions are suboptimal, then we propose \emph{broadcast} as an alternative and show it is provably optimal. Based on this, we develop new algorithms to schedule and load-balance multiple broadcast primitives within a single cross-mesh resharding task. 

\noindent \(\bigstar\) We propose a new pipelining schedule that overlaps multiple cross-mesh resharding communication tasks with the pipeline-parallel computation in a model-parallel job.

\noindent \(\bigstar\) We implement proposed techniques as a communication library, \sys, and show it performs up to 10x faster than existing solutions on various microbenchmarks.

\noindent \(\bigstar\) We integrate the library with Alpa, a state-of-the-art model-parallel system. Our techniques improve the end-to-end multi-node training throughput of two models, GPT and U-Transformer, by 10\% and 50\%, respectively.


\section{Background}
\begin{figure*}[t]
	\centering
	\includegraphics[width=0.9\textwidth]{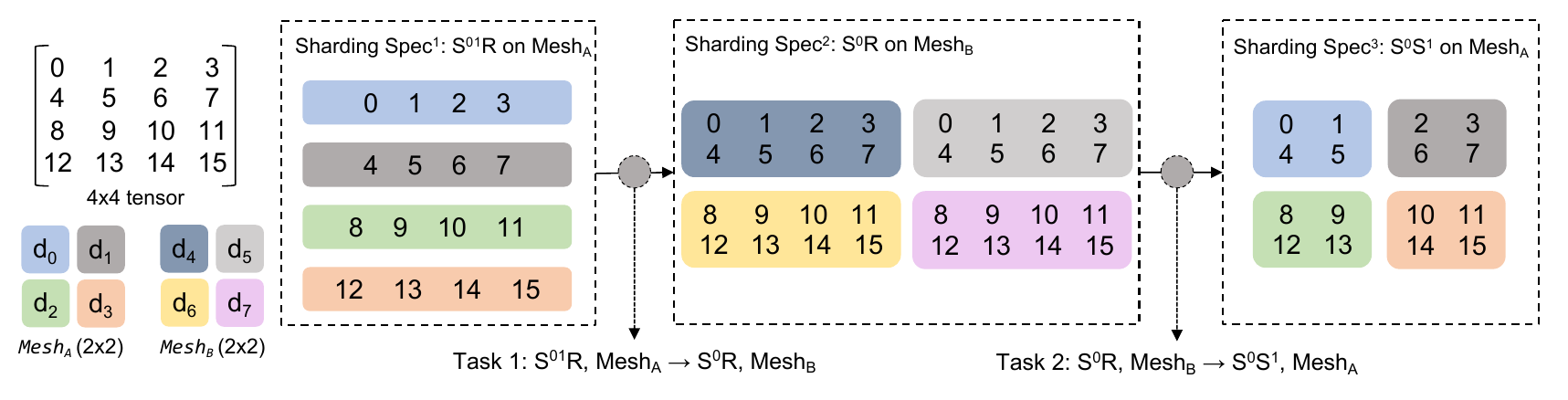}
	\vskip -1em
	\caption{Two examples of cross-mesh resharding. Given a $4 \times 4$ matrix and two $2 \times 2$ meshes $Mesh_A$ and $Mesh_B$, each dotted box shows a sharding spec and its resulting tensor layout on each device of the mesh (note that $Mesh_A$ is used twice: in the first and third dotted boxes). Two cross-mesh resharding tasks are formed when communicating between Specs 1 and 2 and between Spec 2 and 3.}
	\vskip -1.5em
	\label{fig:device-view}
\end{figure*}
In this section, we introduce the basic communication characteristics in model-parallel DL. We then describe a new communication pattern, called \emph{cross-mesh resharding}, that has emerged in large model-parallel workloads.

\subsection{Model Parallelism}
\label{sec:background:mp}

When the model is large and cannot fit into the memory of a single compute device, model parallelism is an essential step to parallelize the model computation over multiple devices. Since DL computation is often defined as a computational graph of tensors and computational operators, model parallelism is equivalent with partitioning the computational graph and placing different parts on parallel devices. Depending on how the graph is partitioned, model parallelism methods can be categorized into two classes: \emph{intra-operator parallelism} and \emph{inter-operator parallelism}, with different communication requirements.

\noindent \textbf{Intra-op parallelism} refers to methods that shard or replicate the operators and its input and output tensors along some tensor axes, and assign different regions of the operator computation to parallel devices, as Figure~\ref{fig:mp-comm}(b) shows.
In practice, intra-op parallelism is realized in an SPMD fashion~\cite{lepikhin2020gshard,xu2021gspmd}: the tensors and computation are \emph{evenly} sharded and dispatched to all participating devices. 
When tensors are sharded along different tensor axes, they may be stored on multiple devices with different \emph{distributed layouts}. When an operator is parallelized using a parallel algorithm, it might require its input tensors to follow a specific layout and produce output tensors with another layout. If an input tensor's  layout disagrees with the required input layout of the operator, a \emph{layout conversion} (also called \emph{resharding}~\cite{xu2021gspmd}) is needed. In intra-op parallelism, this layout conversion is easily facilitated by collective communication operations, such as \emph{all-reduce}, \emph{all-gather}, \emph{all-to-all}, among all participant devices (see Figure~\ref{fig:mp-comm}(b)).

\noindent \textbf{Inter-op parallelism} partitions the graph (instead of individual operators or tensors) and places each resulting subgraph, called \emph{a stage}, onto a device. If an operator's input tensor is generated by another operator on a different device, a point-to-point (P2P) communication is required to send the tensor from the source device to the destination device (see Figure~\ref{fig:mp-comm}(c)). Pipeline parallelism~\cite{huang2019gpipe,narayanan2021memory} is a special case of inter-op parallelism where all devices are arranged into a pipeline and saturated by letting devices from different pipeline stages compute simultaneously in ``assembly-line" fashion; P2P communications are used between stages. Due to data dependencies between partitions, some are idle while waiting its input or the last partition to finish, resulting in bubbles. Bubbles increase if partition is not perfectly even.

\noindent \textbf{Combining Intra- and Inter-op parallelism.}
Neither intra-op parallelism nor inter-op parallelism alone suffices to train large models. Intra-op parallelism always has large communication volume, and cannot scale to multiple nodes, while inter-op parallelism suffers from bubbles. In practice, they must be combined to support large models like GPT-3~\cite{brown2020language}. This combined strategy is implemented in many model-parallel systems~\cite{rasley2020deepspeed, zheng2022alpa, xu2021gspmd} by  first partitioning the computational graph using inter-op parallelism then sharding each stage using intra-op parallelism, shown in Figure~\ref{fig:mp-comm}(d).
Specifically, the graph is first partitioned into multiple stages. Each stage is assigned to a group of devices, referred to as a \emph{device mesh}, sliced from the cluster. Operators and tensors of a stage are parallelized over that stage's assigned mesh following a chosen intra-op parallelism plan; collective communication happens only across devices within each mesh. At the boundary of any two adjacent stages, communication is required to exchange tensors between their meshes. Unlike inter-op parallelism, the tensor might have been sharded with different layouts on the source and destination meshes -- in which case, communication involves not only transferring the tensor, but also performing tensor layout conversion between the source and destination groups of devices. We call this communication pattern \emph{cross-mesh resharding}, which is the focus of this paper. 

In practice, model-parallel systems may sub-optimally map the cross-mesh resharding communication to slower interconnects in a deep learning cluster, such as the Ethernet connection between different nodes/hosts, which has limited communication bandwidth. Such under-optimized mappings can significantly hurt overall training or inference performance (see \S\ref{sec:eval}). 
Next, we formally define cross-mesh resharding and discuss optimization opportunities.


\subsection{Cross-mesh Resharding}
\label{sec:background:cmr}
A model-parallel job may involve many cross-mesh resharding communications. 
We define a \emph{cross-mesh resharding task} as the communication needed to send a single tensor from a source mesh to a destination mesh while meeting the layout requirements. 
To analyze it, we first formalize device mesh and tensor layout.

\noindent \textbf{Device mesh.}
We follow the definitions of mesh and tensor layout in GSPMD~\cite{xu2021gspmd}\footnote{Some model-parallel systems define them differently, which can be equivalently converted to definitions used in this paper.}: a mesh is an n-dimensional array of identical processors. We express a 2D view of a mesh $\mathit{Mesh}_A$ as $(m_1, m_2)$, where $ m_1 \times m_2 = |\mathit{Mesh}_A|$ . For example, a cluster of 2 nodes with 2 GPUs each can be represented as a $(2, 2)$ mesh $[[0, 1], [2, 3]]$, where each integer is a device index, or as a $(1, 4)$ mesh $[[0, 1, 2, 3]]$.


\noindent \textbf{Sharding spec.}
With this definition, we can describe the layout of a sharded $N$-dimensional tensor $D$ over $\mathit{Mesh}_A$ using a \emph{sharding spec}, notated as an $N$-element string: $\mathcal{X} = X_0^{d_0} X_1^{d_1} \dots X_{N-1}^{d_{N-1}}, X_i \in \{S, R\}, d_i \in \{0, 1, 01\}, 0 \le i \le N-1$, 
where $S$ and $R$ stands for \emph{sharded} or \emph{replicated}, respectively. $X_i= S$ means the $i$-th dimension of $D$ is sharded; $X_i = R$ means it is replicated. When a tensor dimension is sharded, a superscript $d_i$ is added and represents which mesh dimensions the sharding are mapped to, e.g., $X_i^{d_i} = S^0$ means the $i$-th dimension of $D$ is sharded along the first dimension of $\mathit{Mesh}_A$; $X_i^{d_i} = S^{01}$ means the $i$-th dim is sharded along both dimensions of $\mathit{Mesh}_A$; $X_i^{d_i} = R$ means that dimension is replicated. Figure~\ref{fig:device-view} illustrates three sharding specs and resulting tensor layouts.

\noindent \textbf{Data slice.}
Due to the sharding or replication, each device in $\mathit{Mesh}_A$ might hold a \emph{data slice} or \emph{replica} of $D$.
For examples, in the 1st sharding spec of Figure~\ref{fig:device-view}, 
each device in $\mathit{Mesh}_A$ holds a unique $4 \times 1$ data slice of $D$;
In the 2nd sharding spec, $D$ is sharded into two unique $2 \times 4$ data slices on $\mathit{Mesh}_B$; devices 4 (blue) and 5 (red) hold a replica of the top unique slice (containing matrix elements 0 through 7).


\noindent \textbf{Cross-mesh resharding.}
With these notations, we can see that a cross-mesh resharding is a \emph{many-to-many multicast communication task} between two non-overlapping device meshes $\mathit{Mesh}_A$, $\mathit{Mesh}_B$ with $\mathit{Mesh}_A \cap \mathit{Mesh}_B = \varnothing$. 
The task aims to send $D$, sharded on the $\mathit{Mesh}_A$ under sharding spec $\mathcal{X}_A$, to $\mathit{Mesh}_B$, where sharding spec of $D$ becomes $\mathcal{X}_B$.  Figure~\ref{fig:device-view} shows two examples of cross-mesh resharding.

It is obvious to see that, regardless of $\mathcal{X}_A$ and $\mathcal{X}_B$, the size of messages transferred between two meshes is lower bound by the size of $D$.
Hence, we can break down a cross-mesh resharding into a set of \emph{unit communication tasks}, and each corresponds to a \emph{unique} data slice $DS_i$ of $D$ on the source $\mathit{Mesh}_A$, and is responsible to send $DS_i$ from $\mathit{Mesh}_A$ to a subset of devices in $\mathit{Mesh}_B$.
For instance, Task 1 in Figure~\ref{fig:device-view} has four unit tasks. The first unit task sends slice $[0, 1, 2, 3]$ to device 4 and 5. 
Task 2 has two unit tasks; its first unit task sends the first $2 \times 4$ slice to device 0 and 1 in $\mathit{Mesh}_A$. In Appendix.\ref{sec:appendix.unit_tasks}, we list all unit tasks of the two cross-mesh resharding in Figure~\ref{fig:device-view}.

Since multiple devices in the source mesh may hold a replica of $DS_i$ (due to replication) and multiple devices in the destination mesh may require $DS_i$, there exist many possible communication routes. Each route may result in different total sizes of messages transferred across the network and traffic on network switches. Moreover, although we have used 2D tensor for explanations, a real model-parallel job works on $N$-dimensional arrays ($N \ge 3$). In \S\ref{sec:single-cmr}, we develop general solutions to minimize the size of message transferred while achieve good load balance for a single cross mesh resharding task. 

Globally, a model-parallel job has many cross-mesh resharding tasks for each tensor exchanged at the boundary of two pipeline stages. Each task is repeatedly performed at every forward and backward pass following the computational dependency of stages. This exposes overlapping opportunities to hide the communication time by scheduling multiple cross-mesh resharding tasks and computation tasks together, as discussed in \S\ref{sec:overlap}.

\section{Optimizations for a single cross-mesh resharding task}
\label{sec:single-cmr}
In \S\ref{sec:background:cmr}, we show that a cross-mesh resharding problem can be decomposed into many unit communication tasks and each unit task is responsible for one data slice. Our goal is to complete all tasks as fast as possible. To achieve that, we frame the original problem as a two-level optimization:
\begin{itemize}
    \item {\bf Optimizing individual unit communication tasks.} We analyze different methods and propose to use broadcast for this part and show that a broadcast-based strategy could give optimal performance. 
    \item {\bf Load balance and schedule for all unit tasks in a cross-mesh resharding.} We formulate the load balance and scheduling problem of multiple unit communication tasks and design two algorithms to solve this problem. Our algorithms can give optimal solutions in most cases. 
\end{itemize}

Before going into details of the two levels, we first consider the typical ML cluster setting with the following properties:
\begin{itemize}
    \item {\bf Fast intra-node and slow inter-node communication.} For example, in a typical GPU cluster, GPUs within a node have fast NVLink interconnect while GPUs across nodes need to communicate via slower Ethernet or Infiniband connections.
    \item {\bf Fully-connected topology between nodes.} We assume that the network bandwidth between a pair of nodes will not be affected by communication between other nodes. In addition, we assume any pair has the same bandwidth. This assumption holds for most clouds and recent datacenters \cite{andreyev2014introducing}.
    \item {\bf Communication bottleneck at hosts.} The bandwidth of a device (e.g., GPU) is often much higher than its host. When multiple devices in a single host send data to anothor host, they will compete for the communication bandwidth at the host's network interface. \yonghao{In this work, we assume each host has only one NIC, which is the common setup in public clouds.}
    \item {\bf Separate sending/receiving bandwidth (full duplex).} We assume each device has separate sending and receiving bandwidth. One device can receive data at full bandwidth while also sending at full bandwidth. This applies to both NVLink and inter-host networks.
\end{itemize}

\begin{figure*}
    \centering
	\begin{subfigure}[t]{.49\columnwidth}
	    \centering
		\includegraphics[width=\textwidth]{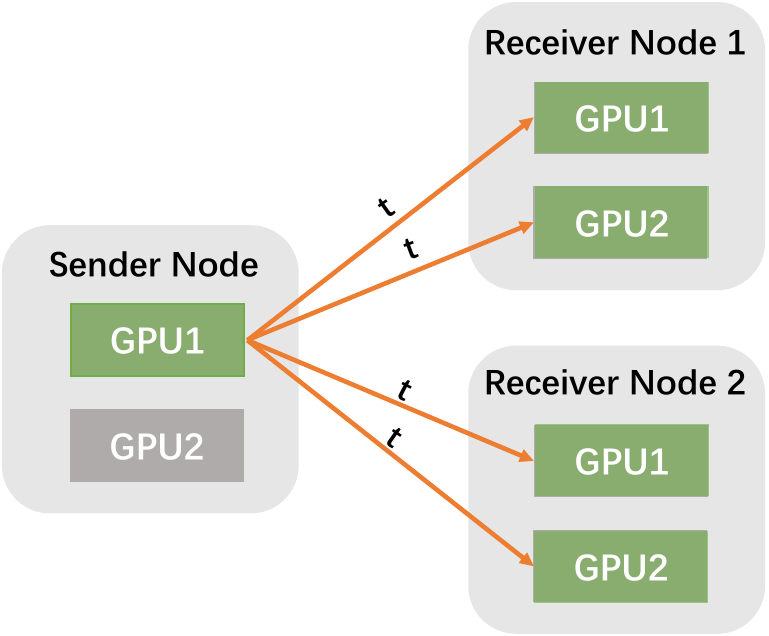}
        \caption{Send/recv.}
        \label{fig:send-recv}
	\end{subfigure}
	\;
	\begin{subfigure}[t]{.49\columnwidth}
	    \centering
		\includegraphics[width=\textwidth]{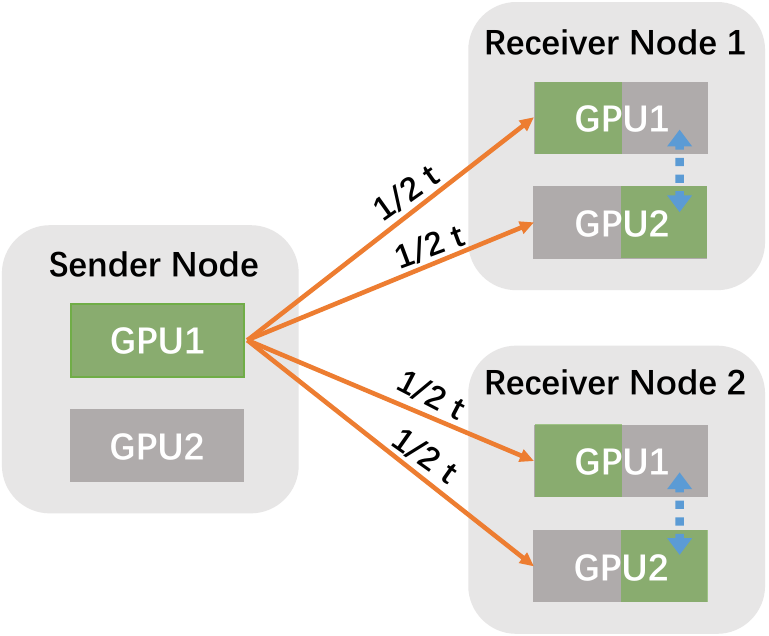}
        \caption{Send/recv + local all-gather.}
        \label{fig:send-recv-local}
	\end{subfigure}
	\;
    \begin{subfigure}[t]{.49\columnwidth}
	    \centering
		\includegraphics[width=\textwidth]{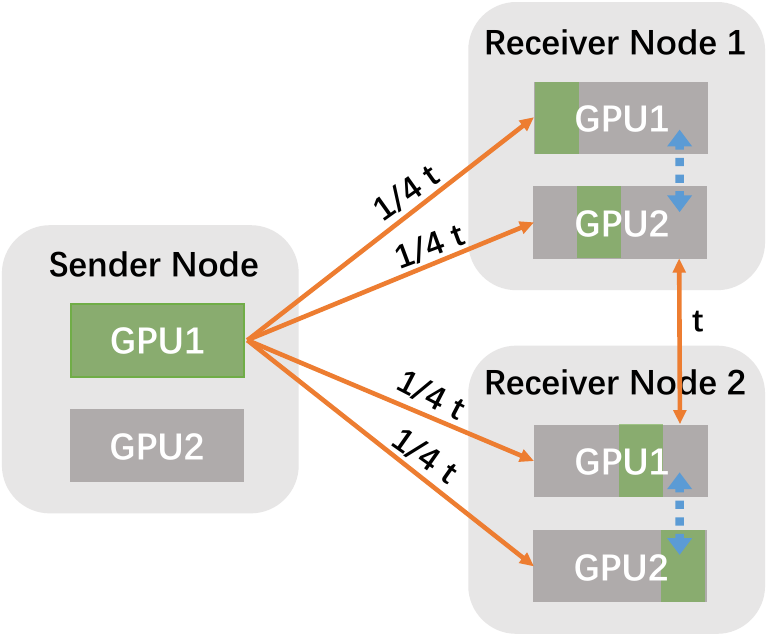}
        \caption{Send/recv + global all-gather.}
        \label{fig:send-recv-global}
	\end{subfigure}
	\;
	\begin{subfigure}[t]{.49\columnwidth}
	    \centering
		\includegraphics[width=\textwidth]{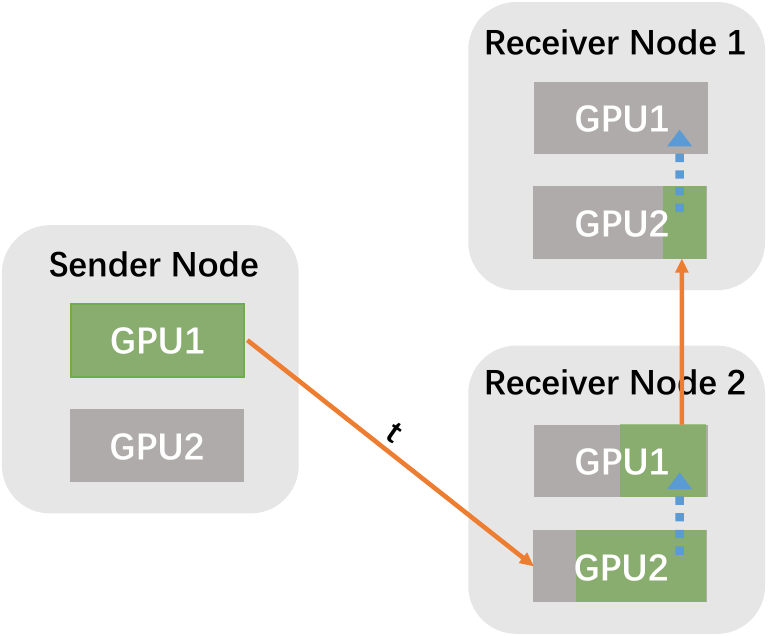}
        \caption{Broadcast.}
        \label{fig:broadcast-based-resharding}
	\end{subfigure}
    \vskip -0.6em
    \caption{Communication strategies for an individual communication task. The orange arrows represent cross-host communication and the numbers on it represent the latency of the communication. The blue arrows represent intra-node communication. 
    }
    \label{fig:comm_strategies}
    \vskip -1.5em
\end{figure*}

\subsection{Optimizing individual unit communication task}
\label{subsec:opt-a-single-comm-task}

We use $\mathit{DS}_i$ to be the data slice corresponding to the $i$-th unit communication task. $\mathit{DS}_i$ has replicas in device set $N_i \subseteq \mathit{Mesh}_A$, and we want to send $\mathit{DS}_i$ to devices in set $M_i\subseteq \mathit{Mesh}_B$. In this subsection, we propose several communication strategies to achieve this goal. 

\yonghao{We first point out that a unit task cannot be directly handled by collective primitives. In a unit task, potential senders does not overlap with receivers, while collective primitives focus on cases where senders are also receivers.}

From the properties of ML clusters described above, we can see that a good communication strategy will (1) minimize inter-host communication volume and (2) maximize the total sending and receiving bandwidth utilization of all participating devices.

In the following analysis, we assume we are sending a fixed-size object that takes duration $t$ to send from a group of nodes to another group and ignore the time we spent on intra-node communication. For a communication strategy, we denote $T(A, B)$ to be the time to send the fixed-size object from one device to $A\times B$ devices with $A$ nodes and $B$ devices on each node. Next, we introduce several strategies, from the most naive to the most effective.

\textbf{Send/recv.} We first consider the case with only one host as the potential sender. The simplest strategy is to let the only sender send the data slice $\mathit{DS}_i$ to each receiver device in $M_i$ one by one. In this case, the total communication latency is 
$T^\mathit{sr}(A,B)=ABt,$
 proportional to the number of total devices, as in Figure~\ref{fig:send-recv}. If some receiver devices are co-located on the same node, this strategy introduces redundant inter-node communication, since the other devices can receive $\mathit{DS}_i$ from peers on the same node that have already received the data slice. When there are multiple potential senders, we can distribute workloads to different senders and speed up the overall performance. We discuss such load balance in the next subsection. 

\textbf{Send/recv with local allgather.} \label{sec:single-resharding-method} One straight-forward improvement to the naive send/recv strategy is to let each receiver node only receive a copy of $\mathit{DS}_i$, and then use the fast intra-node connection to transfer the slices among all devices on this node. To achieve this, we split $\mathit{DS}_i$ into $B$ parts, and send each part to one of the $B$ devices on a specific receiver node. Then the receiver devices on the same node assemble a full copy of $\mathit{DS}_i$ via an \emph{all-gather} collective communication primitive. 
Therefore, the overall latency for a sender is
$T^{\mathit{srla}}(A,B)=At,$ as in Figure~\ref{fig:send-recv-local}.
This strategy improves the naive send/recv by reducing inter-host communication volume, so the latency is only proportional to the total number of hosts. Similar to send/recv, we can distribute the workload with multiple sender candidates. 

\textbf{Send/recv with global allgather.}
We further slice $\mathit{DS}_i$ into more fine-grained chunks. More specifically, we can split $\mathit{DS}_i$ over all $A\times B$ slices, and send one slice to each of the $A\times B$ devices. The latency of this step is $t$ since in total the sender only sends one copy of $\mathit{DS}_i$. These devices then perform a global all-gather to collect all slices for each device, whose latency is also $t$ with a typical ring all-gather algorithm \cite{nccl}. Therefore, the ideal overall latency
$T^{\mathit{srga}}(x,y)=2t,$
which does not grow with the number of nodes or devices, as in Figure~\ref{fig:send-recv-local}.

\textbf{Broadcast based resharding.} 
Both the previous two allgather-based strategies can be decomposed into two stages: the first stage where a sender sends the data slice $\mathit{DS}_i$ to receiver devices, and the second where receiver devices exchange different partitions of the data slice. We observe that we can further optimize this by overlapping these two stages. More specifically, when a receiver starts to receive data from the sender, it can act as another sender to send data to the remaining receiver devices, as illustrated in Figure~\ref{fig:broadcast-based-resharding}. Eventually, this strategy can be viewed as a broadcast from the sender device to all receiver devices.
Suppose we split the data slice $\mathit{DS}_i$ into $K$ partitions. Each device sends the partition to the next device after it received one full partition. The time to send each partition is $t/K$ and the overall latency is
$T^\mathit{bc}(A,B) = \frac{t}{K}\cdot(K+A) = t+\frac{At}{K},$ as in Figure~\ref{fig:broadcast-based-resharding}.
When we pick $K$ to be a relatively large value compared to $A$ (e.g., $K \approx 100$ in our experiments), the overall latency will be around $t,$ which is optimal among all proposed strategies and has already reached the upper bound. $t$ is the communication upper bound because each receiver node has only one NIC, and needs at least one replica, which takes $t$ to receive. The broadcast only involves one sender device from $N_i$ and all receiver devices $M_i$. This is different from previous send/recv-based methods which have to utilize more senders. For the case where receivers have multiple NICs, a unit task can be optimized by dividing it into multiple tasks. We leave this part to our future work.

\subsection{Load balance and schedule for multiple unit communication tasks}
\label{sec:load_balancing}
Each cross-mesh resharding task includes multiple unit communication tasks. These tasks might overlap on both sender and receiver devices and affect each other. Therefore, to optimize the overall completion time for cross-mesh resharding, we treat this problem as a \emph{load balancing and scheduling} problem. Specifically, we need to (1) balance the loads by evenly distributing the communication workloads across sender devices and inter-host communication links to avoid congestion and stragglers; (2) schedule the order of different tasks assigned to specific devices to minimize the wait due to unavailable sender/receiver. 

Since hosts are the main bottleneck for communication, we perform load balancing and scheduling at host level, rather than individual devices. Suppose a unit communication task has replicas $\mathit{DS}_i$ in the host set $n_i \subseteq \mathit{Mesh}_A$ and we want to send $\mathit{DS}_i$ to the hosts in set $m_i \subseteq \mathit{Mesh}_B.$ Our first goal is to pick a host $n_{i*} \in n_i$ to send the data to all the receiver hosts in $m_i$ to balance the loads across different devices. Once we picked the sender host $n_{i*}$ for all tasks, we also need to schedule the execution order of different tasks. Note that all the hosts participate in the $i$-th task need to send or receive data with in a shared period of time and different tasks' execution should not overlap. Suppose the duration of the $i$-th task to be $T_i$ and denote the execution starting time to $S_i$, we can formulate the load balancing and scheduling problem as the following optimization problem:
\begin{align}
    {}& \min_{S, n_*} \max_{i} S_i + T_i; \label{eq:load-balance-objective} \\
    \text{s.t. }{} & n_{i*} \in n_i; \label{eq:load-balance-correct-sender} \\
    & (S_i, S_i + T_i) \cap (S_j, S_j + T_j) = \emptyset, \nonumber \\ 
    & \forall n_{i*} = n_{j*} \text{ or } m_i \cap m_j \ne \emptyset. \label{eq:load-balance-no-overlap} 
\end{align}
The optimization objective (Eq.\ref{eq:load-balance-objective}) minimizes the completion time of the last unit communication task. The two constraints make sure the sender of each task has the correct data slice (Eq.\ref{eq:load-balance-correct-sender}) and the tasks that share the same sender or receiver devices are not overlap with each other (Eq.\ref{eq:load-balance-no-overlap}).\yonghao{With multiple NICs on a node, the formulation still applies by making $n_i$, $n_{i*}$ and $m_i$ represent sets of NICs.}

Next, we introduce several load balancing and scheduling algorithms and analyze their properties. 

\textbf{Naive algorithm.} The most straight-forward algorithm is to assign each task to be sent by the first (i.e., lowest-indexed) device in the sender's host. We schedule all the tasks following an arbitrary global order. We serve this algorithm as our baseline to compare with our later algorithms.
   
\textbf{Load balance only.} We start from the load balance across multiple sender hosts. In this case, the original problem is simplified as
\begin{equation}
\min_{n_*} \max_{k\in \mathit{Mesh}_A} \sum_{i: n_{i*} = k} T_i. \label{eq:load-balance-only-objective}
\end{equation}
We note that this minimax optimization problem can be solved optimally by a classical greedy algorithm: we can sort all tasks in descending order of duration $T_i$. Then, iterate through the sorted tasks and assign each task to the sender host with the currently-lightest workload.

\noindent\textbf{Depth First Search (DFS) with Pruning.} All previous algorithm does not take the scheduling of different tasks into account. We first note that scheduling in the original optimization problem (i.e., deciding $S$ in Eq.\ref{eq:load-balance-objective}-\ref{eq:load-balance-no-overlap}) can be simplified to the following problem: For each host, assign an execution order to all of the send/receive tasks on that host. Then, the starting time of each task $S_i$ can be set to the earliest time at which all preceding tasks have finished on the sender host $n_{i*}$ and the receiver hosts in $m_{i}.$ 

For the complex space of sender assignments and task orders, we first propose a search based algorithm: We construct a depth-first search tree over the two types of decisions. When the algorithm reaches a leaf node with a complete sender assignment and task schedule, we compute the completion time in Eq.\ref{eq:load-balance-objective}. During the search process, we prune the branches whose execution time lower bound (measured by the sum of task durations on a single device as in Eq.\ref{eq:load-balance-only-objective}) exceed the current best completion time.

This algorithm has exponential computation complexity, so in practice, we impose a time budget and return the best solution found within the time budget. However, this algorithm fails to produce an efficient schedule within the time budget when there are $>20$ unit communication tasks. This motivates the next algorithm.

\noindent\textbf{Greedy Search with Randomization.} We propose an iterative greedy algorithm that at each iteration, selects a set of non-overlapping tasks which has the maximum number of devices. To find such a set, we use a randomized algorithm: First, we generate a random ordering over all tasks to be scheduled. Then, we iterate through each task with the generated order and select the tasks that do not overlap with previously selected ones. We repeat this process multiple times and select the largest candidate set for the iteration.

For cross-mesh resharding tasks, since we communicate tensors evenly sharded, many unit communication tasks are identical, in terms of the number of devices involved as well as the amount of transferred data and the tasks are distributed uniformly across different devices. Therefore, the greedy randomized algorithm can find good and even optimal solutions in short amount of time.

\section{Overlapping-friendly Pipeline Schedule}
\label{sec:overlap}
\begin{figure}
	\centering
	\begin{subfigure}[b]{\columnwidth}
      \centering
	   \includegraphics[width=\textwidth]{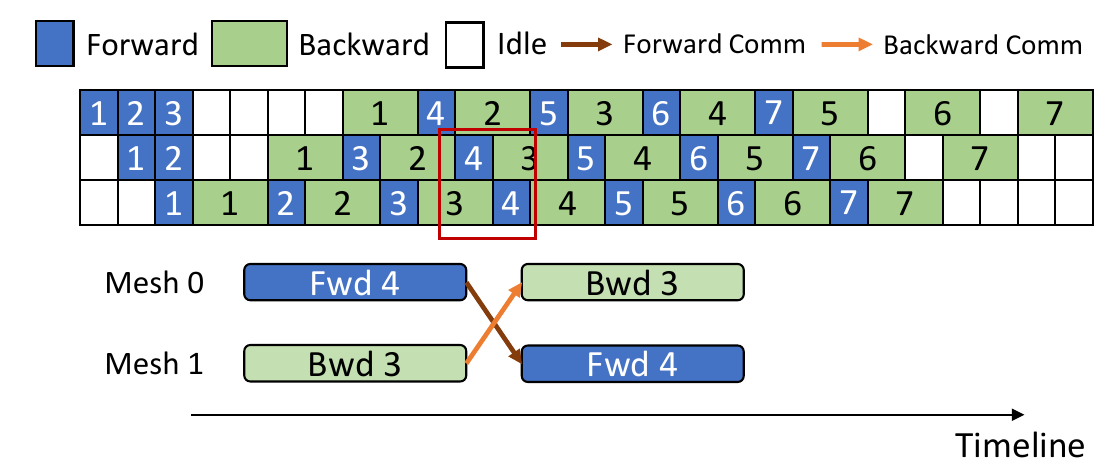}
      \vskip -1em
      \caption{1F1B schedule}
    \end{subfigure}

	\begin{subfigure}[b]{\columnwidth}
	    \centering
		\includegraphics[width=\textwidth]{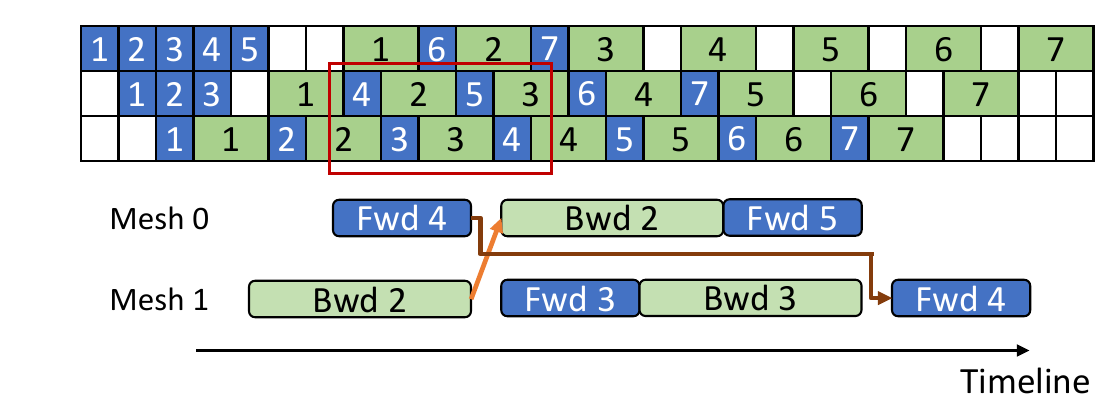}
        \vskip -1.5em
        \caption{Eager 1F1B schedule}
	\end{subfigure}
    \vskip -1em
	\caption{Timelines of 1F1B and eager-1F1B schedule. Each row is a pipeline stage. Numbers indicate micro-batch indices. Micro-batch 4's communication from stage 2 to stage 3 is magnified.}
	\label{fig:overlap:schedules}
    \vskip -1em
\end{figure}

\begin{table}[t]
\centering
\caption{Size of parameters, optimizer states, and activations per GPU for a GPT-3 layer in mixed precision training. The sequence length $S = 1024$, hidden size $H = 12288$, per-GPU micro-batch size $B = 2$, and tensor model parallel degree $\text{TMP} = 8$.}
\vskip 0.2em
\scalebox{0.85}{
\begin{tabular}{c|c|c}
\toprule
 &  expression & value \\
\midrule
\#parameter             & $12H^2/\text{TMP}$   & 216M\\
\#optimizer state parameters  & $24H^2/\text{TMP}$   & 432M\\
\#activation elements        & $BSH$         & 24M\\
Memory of weights and optimizer & $168H^2/\text{TMP}$   & 2.95GB\\
Memory of activation   & $2BSH$    & 48MB\\
\bottomrule
\end{tabular}
}
\label{table:overlap:model-mem}
\vskip -1.0em
\end{table}

The previous section addresses how to optimize a single cross-mesh resharding task. However, this is not enough for performing pipeline parallelism which requires scheduling multiple compute and communication tasks from multiple micro-batches and stages.
To study the overhead of cross-mesh communication, we implement a hypothetical upper bound called "Signal Send/Recv". This method only communicates one byte for all cross-mesh resharding tasks, so it keeps the data dependency of the compute tasks but removes almost all communication costs. We find the training throughput can be 1.1x to 1.5x faster with "Signal Send/Recv" (\S\ref{fig:eval:e2e}) . This means there is still an opportunity to reduce communication costs.

Our key observation is that the existing synchronous pipeline schedules such as 1F1B~\cite{narayanan2021efficient} and GPipe~\cite{huang2019gpipe} are not overlapping-friendly. Fig.~\ref{fig:overlap:schedules}(a) shows the timeline of 1F1B schedules.
The following three tasks of the same micro-batch are scheduled contiguously: the computation of $i$-th stage, the communication between $i$-th stage and $(i+1)$-th stage, the computation of $(i+1)$-th stage. Due to this strict data dependency, the communication cannot be overlapped.

To solve the problem, we develop a novel and more overlap-friendly schedule called eager-1F1B. The main idea is to run computations eagerly to create opportunities for overlapping. Fig.~\ref{fig:overlap:schedules}(b) shows the timeline of the new schedule.

The eager-1F1B schedule is based on 1F1B but shifts forward computation tasks to several time steps ahead.
During the warm-up phase of the original 1F1B, stage $i$ runs forward computations for the first $(\#stages - i + 1)$ batches. Stage $i$ then stops for a while and enters the steady one-forward-one-backward phase after it is able to run the first backward computation.
During the warm-up phase of the eager-1F1B schedule, stage $i$ runs forward computations for more micro-batches. Stage $i$ runs $(2\times(\#stages - i) + 1)$ forward computations and then enters the steady phase.

When there is no communication cost, the latencies of these two schedules are the same. However, when the communication cost is not negligible, 1F1B cannot hide the cost due to the strict data dependency.
On the contrary, the eager-1F1B schedule inserts computation tasks from other micro-batches between two dependent tasks, making it possible to hide the communication costs. One example of this overlapping is shown as the magnified plot in Fig.~\ref{fig:overlap:schedules}.

The downside of the eager-1F1B schedule is a slight memory usage increase because each GPU stores temporary activation tensors for more micro-batches. 
However, this increment is very small in common cases.
We list the size of parameters and activations with a common parallel configuration in Table~\ref{table:overlap:model-mem}.
The new schedule increases per-GPU memory usage by at most $\#stages\times size_{{activation}}$, which is a lot smaller than the parameter size.

The eager-1F1B schedule can overlap the communication of the forward part but does not help the backward part.
To overlap backward communication, we divide a backward computation into two parts: computing gradients of activations and gradients of weights.
The first does not rely on the second, and cross-mesh communication only uses the output of the first. Hence, we delay the second part so that it overlaps with the communication. We call this backward weight delaying.
Similarly, this technique slightly increases the peak memory usage, so we use a simple cost model to estimate the compute and communication time and delay the least to cover all communications.

We find that in practice, even without backward weight delaying, the performance of the eager-1F1B schedule is very close to the "Signal Send/Recv". This is because later pipeline stages store fewer activations and have less memory pressure. As a result, they use less rematerialization and are slightly faster, which provides time for communication.

\section{Evaluation}
\label{sec:eval}

The optimizations in this paper are implemented with about 900 lines of C++ and 2.5k lines of python.
We implement a standalone cross-mesh resharding communication library \sys for GPU based on NCCL~\cite{nccl} and integrate it into Alpa~\cite{zheng2022alpa}, a state-of-the-art distributed training framework. We implement the eager-1F1B schedule and overlap optimizations on top of Alpa, which provides APIs to customize pipeline schedules and cross-pipeline-stage communication. Implementing in platforms like PyTorch is possible, but we believe it needs more effort.

We first evaluate the communication library on a set of microbenchmarks of cross-mesh resharding cases. We then evaluate all optimizations on end-to-end training of large models with billions of parameters, including a \mbox{GPT-3} like language model~\cite{brown2020language} and a U-Net Transformer~\cite{petit2021unettransformer}. We also isolate each part in our optimizations and perform ablation studies of them.

We run experiments on an AWS cluster where each node is a p3.8xlarge instance with 4 NVIDIA V100 (16GB) GPUs and 32 vCPUs. GPUs in a node are connected via NVLink and nodes are launched with 10Gbps cross-node bandwidth.

\subsection{Microbenchmark}
\label{sec:eval:microbenchmark}

We first show the advantage of our broadcast-based resharding when sending the tensor from a single device to multiple devices. We then show the advantage of broadcast-based resharding with load balance when sending the tensor from multiple devices to multiple devices.

\begin{figure}
	\centering
    \vspace{-0.5em}
	\begin{subfigure}[b]{.5\columnwidth}
      \centering
	   \includegraphics[width=\textwidth]{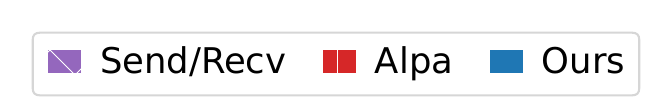}
    \end{subfigure}
	\begin{subfigure}[b]{.49\columnwidth}
	    \centering
		\includegraphics[width=\textwidth]{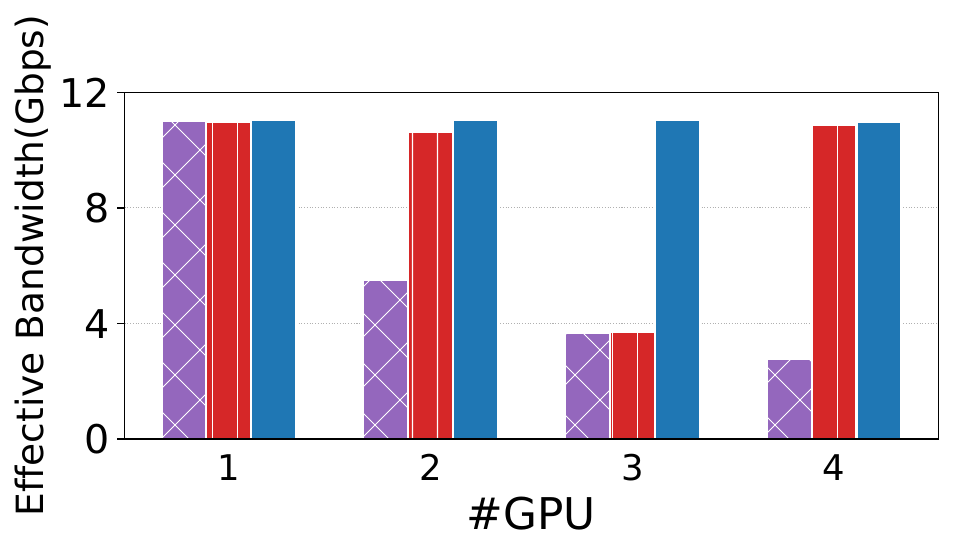}
        \vskip -0.75em
        \caption{one receiver node.}
	\end{subfigure}
	\begin{subfigure}[b]{.49\columnwidth}
	    \centering
		\includegraphics[width=\textwidth]{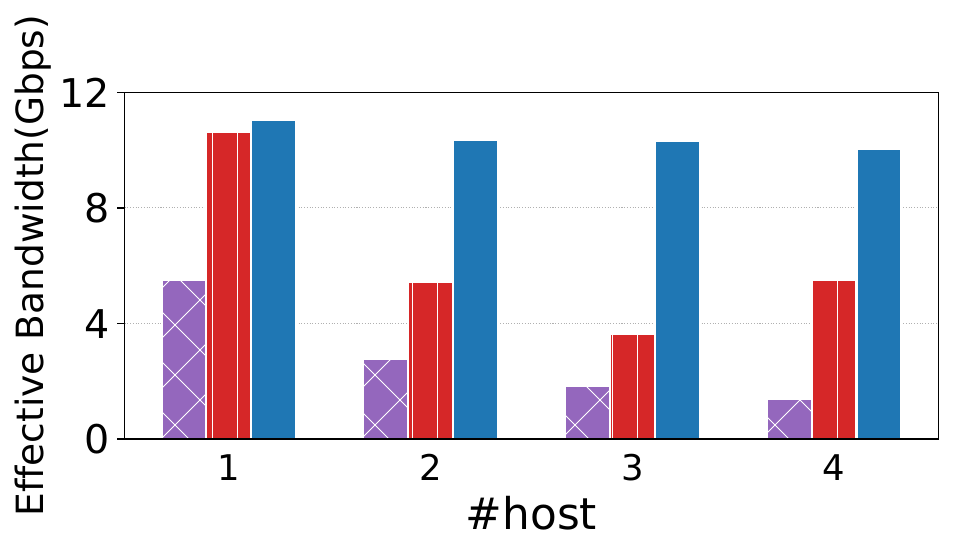}
        \vskip -0.75em
        \caption{multi receiver nodes.}
	\end{subfigure}
    \vskip -1em
	\caption{Single device to multiple devices microbenchmark result.}
	\label{fig:eval:microbenchmark-single-sender}
    \vspace{-1.5em}
\end{figure}

\subsubsection{Single device to multiple devices}
\label{subsubsec:single-to-multiple}

\textbf{Experimental setup.}
In this setting, the sender mesh has only 1 GPU. We vary the number of GPUs in the receiver mesh. In the first group of benchmarks, the device mesh has 1 node and the number of GPUs in this node varies from 1 to 4. In the second group of benchmarks, the number of GPUs per node is fixed at 2, but the number of nodes grows from 1 to 4.
Both the sender and receiver use a fully replicated sharding spec.
The message size keeps 1GB.

\textbf{Baseline.}
We compare our work against two baselines: "Send/Recv" and "Alpa".
"Send/Recv" only uses point-to-point send/recv communication primitives as in \ref{sec:single-resharding-method}.
"Alpa" is an all-gather based approach used in systems like Alpa and Megatron-LM, which can offload communications from slow cross-node connections to fast intra-node connections. 

\textbf{Results.}
As Fig.\ref{fig:eval:microbenchmark-single-sender} shows, when the receiver mesh has only one node, the latency of “Send/Recv" grows linearly with the number of GPUs in the mesh, while ours \sys and Alpa only have less than 1\% increase. The result is aligned with our analysis for broadcast and allgather based method(Alpa) in \ref{subsec:opt-a-single-comm-task}. 
However, when the number of nodes in the receiver mesh is greater than 1, the intra-mesh all-gather in Alpa has to use the inter-node connections as well, and hence the overhead of all-gather is no longer negligible. On the contrary, \sys still shows almost no performance degradation.
The sudden performance drop of Alpa when \#gpu is 3 or \#node is 3 is because Alpa cannot handle uneven partition, while ours \sys efficiently handles tiling, padding, and pipelining with the broadcast-based approach.

\subsubsection{Multiple devices to multiple devices}
\label{sec:eval:microbenchmark-multi-device-sender}
\begin{table}[t]
\centering
\caption{Sharding specs of senders and receivers in multi-to-multi-device microbenchmark.}
\vskip 0.5em
\scalebox{0.72}{
\begin{tabular}{c|cccccc}
\toprule
case & send spec & recv spec & send mesh shape & recv mesh shape\\
\midrule
case1 & $S^0RR$ & $S^0RR$ & (2.4) & (2,4)\\
case2 & $RRR$ & $S^0RR$ & (2,4) & (2,4)\\
case3 & $RS^0R$ & $S^0RR$ & (2,4) & (2,4)\\
case4 & $RS^{01}R$ & $S^{01}RR$ & (2,4) & (2,4)\\
case5 & $S^1RR$ & $S_{0}RR$ & (2,4) & (2,4)\\
case6 & $S^0RR$ & $S^0RR$ & (2,4) & (3,4)\\
case7 & $S^1RR$ & $RRR$ & (1,4) & (2,4)\\
case8 & $RRR$ & $RRR$ & (2,3) & (3,2)\\
case9 & $RS^0R$ & $RRS^0$ & (2,4) & (2,4)\\

\bottomrule
\end{tabular}
}
\label{table:eval:microbenchmark-sharding-spec-configs}
\end{table}
\begin{figure}
    \centering
    \vspace{-0.5em}
    \includegraphics[width=0.95\columnwidth]{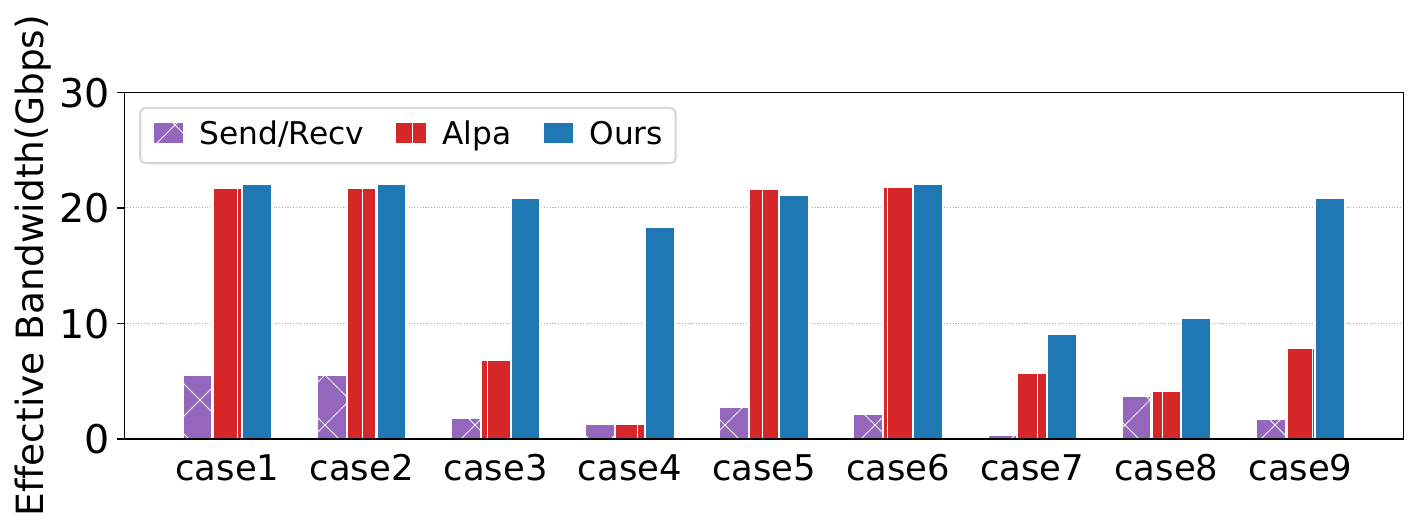}
    \vskip -1em
    \caption{Multi-to-multi-device microbenchmark result.}
    \label{fig:eval:microbenchmark-multi-sender}
    \vskip -1.5em
\end{figure}

\textbf{Experimental setup.}
Next, we move to more complicated cases where both the sender mesh and receiver mesh have multiple nodes.
Table~\ref{table:eval:microbenchmark-sharding-spec-configs} lists the configurations of all cases.
We use a tensor of shape (1024, 1024, 512) and pad it in case 6.
We pick several representative sharding specs from common deep learning workloads. For example, sharding spec $S^0RR$ is the spec of activation tensors in transformer models when using combined data and operator parallelism;
$S^{01}RR$ is the spec when using pure data parallelism; $RRR$ is the spec when using pure operator parallelism.
We use the same set of baselines as in \S\ref{subsubsec:single-to-multiple}. The baselines do load balancing with a greedy approach which picks the sender with the lowest load for the next data slice, as in \ref{sec:load_balancing}.

\textbf{Results.}
In case 1, 2, 5, and 6, \sys and Alpa perform similarly. Both benefit from offloading communications to fast NVLinks. In case 7 and 8, Alpa's all-gather crosses nodes and is slow, but \sys pipelines it with the inter-mesh communication, so it is up to 2.5x faster than Alpa.
In case 3, 4, and 9, \sys is 3x to 10x faster than Alpa respectively, because it reorders communications to utilize bandwidth of both two nodes of sender.
In contrast, with Alpa's order, two sender nodes always share the same receiver, making one of them idle. This problem is more significant when the number of tiles is large, so \sys shows an even better speedup in case 4 which schedules 64 unit tasks. 

\subsection{End-to-End Performance}
\begin{table}[t]
\centering
\caption{Models in end-to-end evaluation. Parallel config is a tuple of (data parallel degree, operator parallel degree, pipeline parallel degree).}
\scalebox{0.8}{
\begin{tabular}{ccccc}
\toprule
Model & Batch Size & \#params & Precision & Parallel Config \\
\midrule
GPT case1 & 1024 & 1.3B & FP16 & (2, 2, 2) \\
GPT case1 & 1024 & 2.6B & FP16 & (2, 2, 2) \\
GPT case2 & 1024 & 2.6B & FP16 & (4, 1, 2) \\
U-Trans case1 & 2048 & 1B & FP16 & (auto, auto, 2) \\
U-Trans case2 & 2048 & 2.1B & FP16 & (auto, auto, 2) \\
U-Trans case3 & 2048 & 2.1B & FP32 & (auto, auto, 2) \\
\bottomrule
\end{tabular}
}
\vskip -0.5em
\label{table:eval:model-config}
\end{table}

\begin{figure}
	\centering
	\begin{subfigure}[b]{\columnwidth}
      \centering
	   \includegraphics[width=\textwidth]{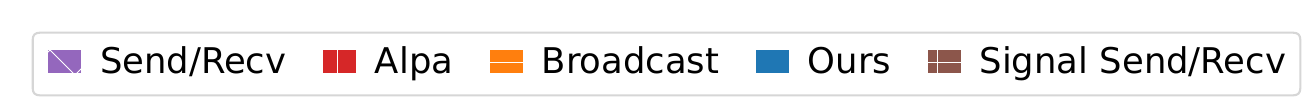}
    \end{subfigure}

	\begin{subfigure}[b]{.56\columnwidth}
	    \centering
		\includegraphics[width=\textwidth]{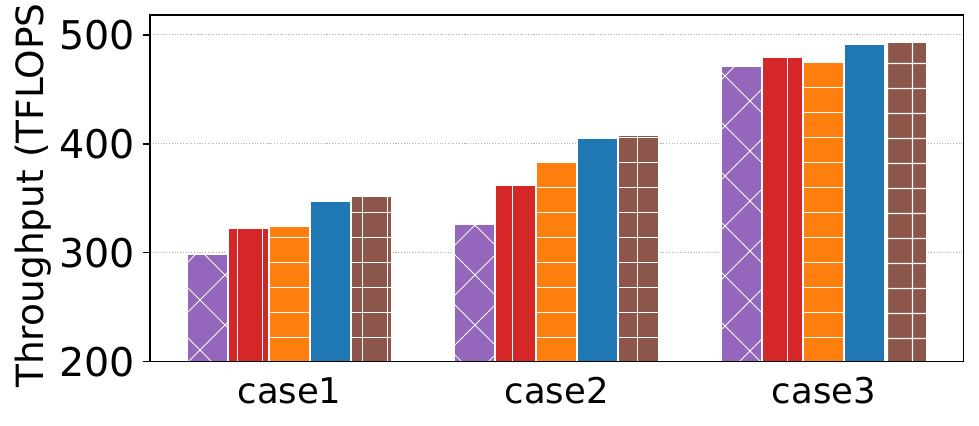}
        \caption{GPT}
	\end{subfigure}
	\begin{subfigure}[b]{.4\columnwidth}
	    \centering
		\includegraphics[width=0.62\textwidth]{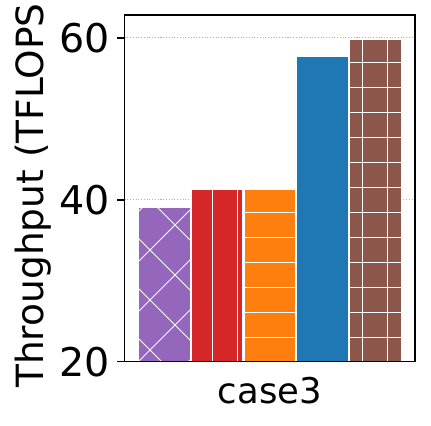}
        \caption{U-Trans, FP32}
	\end{subfigure}

    \vspace{0.5em}
    \begin{subfigure}[b]{.5\columnwidth}
	    \centering
		\includegraphics[width=0.8\textwidth]{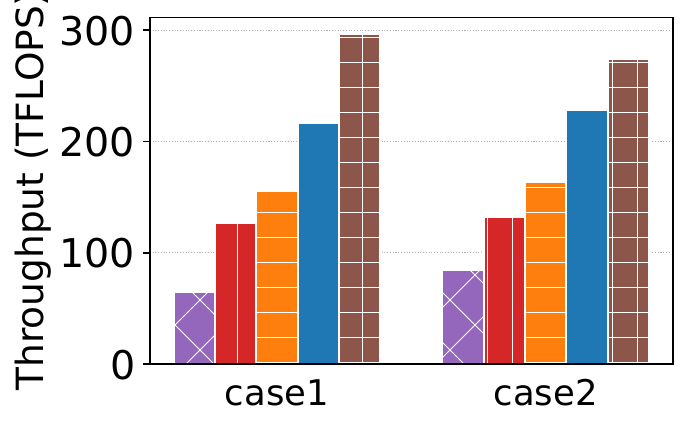}
        \caption{U-Trans, FP16}
	\end{subfigure}
    \vskip -1em
	\caption{End-to-end evaluation result.}
    \vskip -2em
	\label{fig:eval:e2e}
\end{figure}

\textbf{Experiment setup.} We evaluate our work on two models of which the configuration is listed in Table~\ref{table:eval:model-config}.
GPT is a homogeneous transformer-based model built by stacking transformer layers with the same structure. At each pipeline stage, it sends the output activation tensor of the last transformer layer in the stage.
The tensor is partitioned on data-parallelized devices and replicated on operator-parallelized devices.
U-Transformer~\cite{petit2021unettransformer} is a U-shaped convolution neural network with long skip connections. 
It inserts attention layers after convolution layers. In the U-Net model, the i-th layer sends its output activation to not only the next layer, but also the i-th layer from last.
For the GPT model, we test two parallel configurations which combine data, operator and pipeline parallelism. For the U-Transformer model, we manually partition it into two stages and use Alpa to generate the intra-operator parallel strategy of each stage. We balance pipeline stages with respect to FLOPs.

\textbf{Baselines.} We compare our work against three baselines.
The first two "Send/Recv" and "Alpa" are the same as that in \S\ref{subsubsec:single-to-multiple}.
The third is "Broadcast", which uses broadcast-based resharding without overlapping introduced in \S\ref{sec:overlap}.
Therefore, "Broadcast" only optimizes a single resharding task, which mimics streaming optimization introduced in related work such as CoCoNet~\cite{jangda2022coconet}.
We also add the performance of "Signal Send/Recv" as the upper bound, which is discussed in \S\ref{sec:overlap}.

\textbf{Results.}
Fig.~\ref{fig:eval:e2e} shows the evaluation results. The y-axis is the aggregated throughput of all GPUs in the cluster. We use TFLOPs as the throughput metric following the existing literature~\cite{narayanan2021efficient}.

On GPT models, both \sys and Alpa are very close to the upper bound "Signal Send/Recv", because they can offload communication to high-bandwidth NVLink. \sys shows a 1.1x speedup because of overlapping.
On the U-Transformer model, the U-shaped skip connection makes cross-mesh communication a bottleneck. Our eager 1F1B schedule can effectively overlap the communication and achieve a 1.5x speedup compared to Alpa. \sys reaches more than 75\% of the upper bound in all cases.

The communication exhibits a sublinear increase in relation to the growth of models, whereas the computation linearly or even superlinearly increases. This disparity yields a relatively smaller proportion of communication, and thus enables better overlapping and brings \sys closer to Signal Send/Recv. A similar effect can be observed when transitioning from FP16 to FP32 in U-Transformer.

\subsection{Ablation Study}
In this section, we study the effectiveness of our load-balance algorithm and overlap-friendly schedule.
\subsubsection{Load balance}
\begin{figure}
	\centering
    \vskip -0.5em
	\includegraphics[width=0.95\columnwidth]{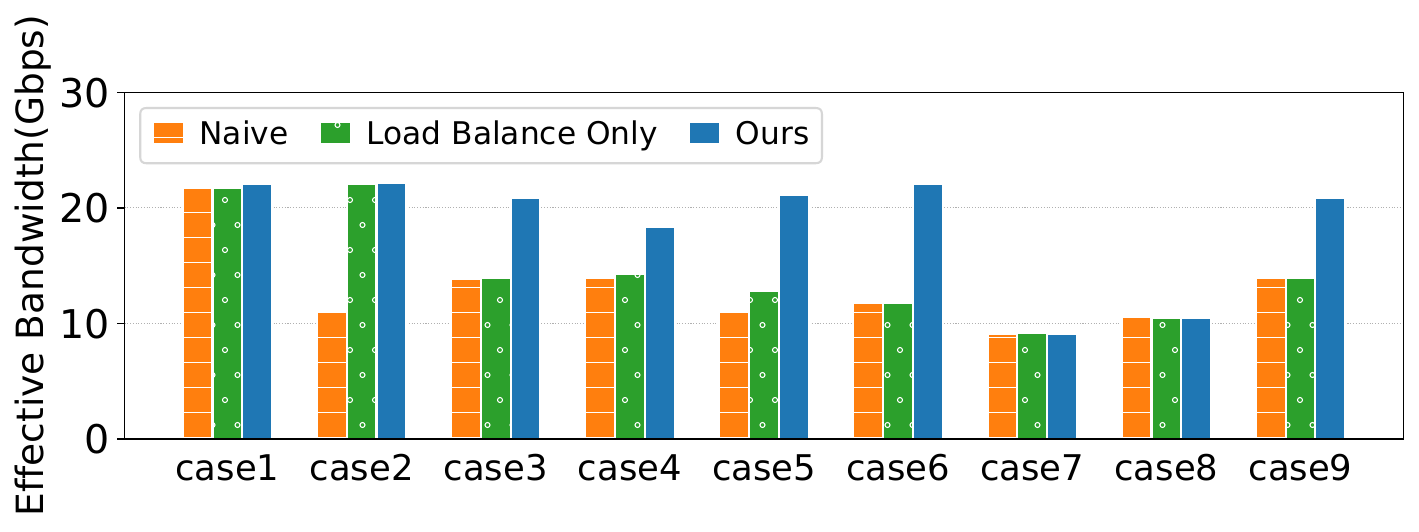}
    \vskip -1.5em
	\caption{Ablation study for load balance algorithm.}
    \vskip -1.5em
	\label{fig:eval:ablation-balance}
\end{figure}
\textbf{Experiment setup.} We use microbenchmarks for the ablation study of load balance. The setup shares \S\ref{sec:eval:microbenchmark-multi-device-sender}'s.

\textbf{Baselines.} We compare our load balance algorithm with two baselines. "Naive algorithm" just chooses the sender with lowest index as in \S\ref{sec:load_balancing} and sends to all receivers for each tile, and sends tiles by an arbitrary global order. "Load balance only" balances the size sent by each node and uses order generated by the greedy algorithm in \S\ref{sec:load_balancing}. \sys is the ensemble of DFS with pruning and the Greedy Search with Randomization. We run both algorithms and choose the better result as the schedule to launch broadcasts.

\textbf{Results.} There is only point-to-point communication in case 1 and 8, so all methods perform similarly. In case 2, "Naive algorithm" sends all tensors from the first node, leading to congestion. In case 8, there is only one broadcast to be launched and thus we have no opportunity to optimize order. In other cases, both "Naive algorithm" and "Load balance only" meet congestion, while \sys can find the optimal order to avoid any send or receive node being idle.

\subsubsection{Overlap}
\begin{figure}
	\centering
	\includegraphics[width=0.7\columnwidth]{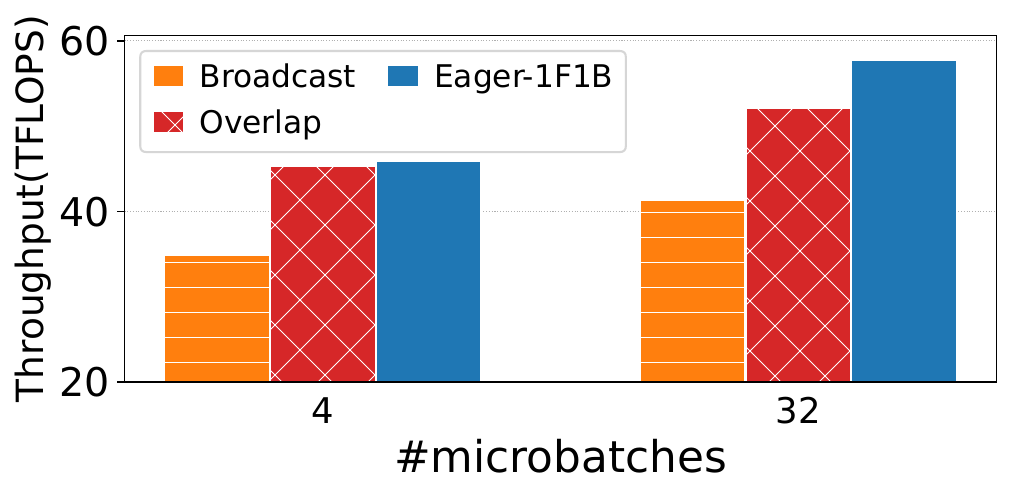}
    \vskip -1.5em
	\caption{Ablation study for overlap-friendly schedule.}
	\label{fig:eval:ablation-overlap}
    \vskip -1.8em
\end{figure}
\textbf{Experiment setup.} We report the performance of our schedule and other variants on U-Transformer model. We use two batch sizes and keep the same microbatch size.

\textbf{Baselines.} We compare our work with two baselines.
"Broadcast" uses broadcast-based resharding and load balance.
"Overlap" overlaps pipeline communication and computation on top of Broadcast, but does not use the eager-1F1B schedule.
Our work "Eager-1F1B" uses the eager-1F1B schedule on top of "Overlap".

\textbf{Results.} For a very small number of microbatches, the pipeline has no steady period, so "Overlap" is only 3\% slower than \sys. For a more typical case with a larger number of microbatches, "Overlap" is 30\% faster than "Broadcast", and "Eager-1F1B" adds another 15\%.


\section{Related Work}

\textbf{Distributed machine learning}
The training of neural networks can be distributed with various parallel methods.
Data parallelism distributes the training by partitioning the input data.
Horovod~\cite{sergeev2018horovod} and PyTorchDDP~\cite{li2020pytorchddp} are two widely adopted data-parallel systems.
ZeRO~\cite{rajbhandari2020zero, rajbhandari2021zero} reduces the memory usage of data parallelism for training large models.
On the other hand, model parallelism distributes the training by partitioning the model, which can be performed at either the inter-operator level or intra-operator level.
Mesh-TensorFlow~\cite{shazeer2018mesh}, GSPMD~\cite{xu2021gspmd}, and GPipe~\cite{huang2019gpipe} provide annotation APIs for users to manually specify a parallel plan.
Megatron-LM~\cite{shoeybi2019megatron, narayanan2021efficient, korthikanti2022reducing} and TeraPipe~\cite{li2021terapipe} design specialized partition strategies for transformer models.
Besides manually designing the parallel methods, Alpa~\cite{zheng2022alpa}, Unity~\cite{unger2022unity}, FlexFlow~\cite{jia2018beyond}, and AutoSync~\cite{zhang2020autosync} automatically searches for the best plan.
The optimizations introduced in this paper can be easily integrated into the above systems to address the communication bottleneck in distributed training.

\textbf{Collective communication}
Most communication in distributed training can be implemented by collective communication primitives.
Efficient algorithms and implementations of these primitives are a long-standing research topic~\cite{chan2007collective, nccl}.
Earlier work solves the problem for regular topology~\cite{scott1991efficient, barnett1994interprocessor,patarasuk2009bandwidth}. Recently, more algorithms are proposed to exploit the hierarchy, heterogeneousness and dynamics~\cite{cho2019blueconnect,wang2020blink,luo2020plink,zhuang2021hoplite}.
The optimal algorithm of these primitives on a specific topology can be synthesized using constraint solver, type-directed or syntax-guided approaches~\cite{cai2021synthesizing, shah2021synthesizing, xie2022synthesizing, rink2021memory}.
On the other hand, domain-specific languages and compilers are built to allow manual optimization~\cite{jangda2022breaking, cowan2022msccl}.
The collective communication works well for the single program, multiple data (SPMD) pattern. However, when doing model parallelism with multiple device meshes, new patterns such as multiple program multiple data (MPMD) and cross-mesh communication appear~\cite{barham2022pathways}. Existing work does not address the new patterns while our paper first comprehensively studies them.

\textbf{Overlapping of computation and communication}
Overlapping is an effective technique to hide the latency of communication.
Prior work has studied scheduling and partition methods to achieve better overlapping \cite{hashemi2019tictac,jayarajan2019priority,jiang2020unified,jangda2022breaking}. They either focus on data parallelism or low-level communication primitives, while this paper proposes a new high-level pipeline schedule for model parallelism. 

\section{Conclusion}
In this paper, we introduce communication optimizations to accelerate the cross-mesh resharding, which is a new and important communication pattern in distributed training of large-scale neural networks. 
We optimize the problem at three granularity including a unit communication task, the load balance among all unit tasks in a cross-mesh resharding task and overlapping all cross-mesh resharding tasks with computations. We port a state-of-the-art distributed training framework, Alpa, on top of our work. Our work speeds up training of large models by up to 1.5x and achieves a performance very close to the hypothetical upper bound.

\bibliography{ref}
\bibliographystyle{mlsys2023}


\newpage
\appendix
%
%
%
%
%





\section{Artifact Appendix}


\subsection{Artifact check-list (meta-information)}

{\small
\begin{itemize}
  \item {\bf Compilation: }Compile the jaxlib binary, script provided;
  \item {\bf Binary: }compiled jaxlib binary;
  \item {\bf Run-time environment: }CUDA toolkit 11.1 or higher;
  \item {\bf Hardware: }multiple GPU with hierarchical bandwidth;
  \item {\bf Metrics: }Throughput;
  \item {\bf Output: }Execution time and throughput;
  \item {\bf How much disk space required (approximately)?: }up to 20 GB;
  \item {\bf How much time is needed to prepare workflow (approximately)?: }up to 1 hour;
  \item {\bf How much time is needed to complete experiments (approximately)?: }about 1 hour;
  \item {\bf Publicly available?: }Merged to Alpa main branch online.
  \item {\bf Code licenses (if publicly available)?: }Apache-2.0
  \item {\bf Workflow framework used?: }Alpa, Flax
  \item {\bf Archived (provide DOI)?: }Available online
\end{itemize}

\subsection{Description}

\subsubsection{How delivered}
We implement our method on top of Alpa, and our code is already merged into its main branch.

\subsubsection{Hardware dependencies}
Our work depends on the GPU environment, and the configuration of our experiment is in table \ref{table:artifact_eval_configs}

\begin{table}[h]
\caption{hardware configurations}
\label{table:artifact_eval_configs}
\begin{tabular}{c|c}
    \hline
    Hardware & Requirements \\
    \hline
    Instance number & 2 to 5\\
    \hline
    GPU & 4 Tesla V100 16 GB \\
    \hline
    GPU peer-to-peer & NVLINK \\
    \hline
    Processor & \makecell[c]{AWS custom Intel(R) Xeon(R)\\ Scalable (Skylake) vCPUs\\}\\
    \hline
    Memory & 128 GB \\
    \hline
    Network bandwidth & 10 Gbps\\
    \hline
\end{tabular}
\end{table}

\subsubsection{Software dependencies}

The reader is supposed to install git and python3 with pip.

The user is also supposed to install nightly alpa and its prerequisites including cupy, nccl, and some python libs. The tutorial is available in \url{https://alpa.ai/install.html#method-2-install-from-source}.

\subsection{Installation}
We create a branch for the ease of artifact evaluation: \url{https://github.com/alpa-projects/alpa/tree/mlsys23-artifact}. To run our nightly code, the user is supposed to:
\begin{enumerate}
\item clone the code above;
\item compile our specified jaxlib following the instruction of alpa's installation guide;
\item install the compiled jaxlib under "build\_jaxlib/dist" with the correct jax version under "third\_party/jax";
\item install alpa under the cloned folder.
\end{enumerate}

In "artifact\_evaluation\_scripts" folder, we provide a script to set up the environment("setup\_env.sh").
\subsection{Experiment workflow}

We provide a script that first runs experiments in our micro-benchmark one by one, then executes the end-to-end benchmark: "run\_benchmark.sh". The user can also use "microbenchmark.py" or "benchmark.py" to run specific benchmark cases.
\subsection{Evaluation and expected result}

The evaluation results will be shown on screen during the experiment workflow. For microbenchmarks, the result will be recorded in a json file of each test suite. For end-to-end experiments, the result will be recorded in a tsv file.

The user is expected to have the same result shown in our paper: in micro-benchmarks, resharding\_mode="broadcast" is supposed to outperform resharding\_mode="send\_recv", whether use local allgather or not; For the "resharding\_loadbalance\_mode" term, "loadbalance\_order" is supposed to outperform "loadbalance\_size", which is better than "no\_loadbalance". In end-to-end benchmarks, overlapping with specific pipeline schedule is better than overlapping without the schedule, which is better on the u-net than turning off the overlapping.
\subsection{Experiment customization}

There are some other configurations for users with different numbers of GPUs in the "suite\_unet.py" and "suite\_manual\_gpt.py" files. The user can also select which case of the microbenchmark to run by configuring the "suite\_microbenchmark.py" file.

Though we run our microbenchmark assuming the sender and receiver are one different node, we provide an alternative to let them share the same node by adding --functional-test.






\section{Decomposing to Unit Tasks}
\label{sec:appendix.unit_tasks}
\subsection{Example of all unit tasks from a cross-mesh resharding}
\begin{figure}
	\centering
	\includegraphics[width=0.95\columnwidth]{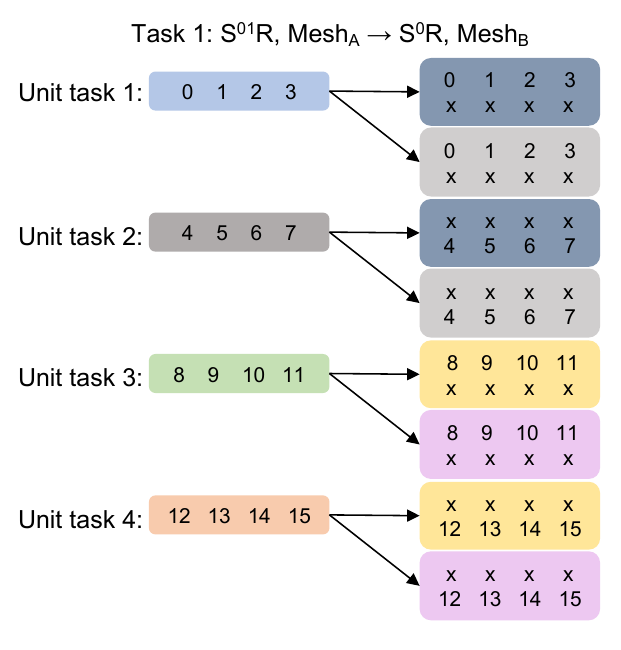}
    \vskip -1.5em
	\caption{Unit tasks for cross-mesh resharding case 1.}
    \vskip -1.5em
	\label{fig:unit_task_1}
\end{figure}
\begin{figure}
    \vskip 1em
	\centering
	\includegraphics[width=0.9\columnwidth]{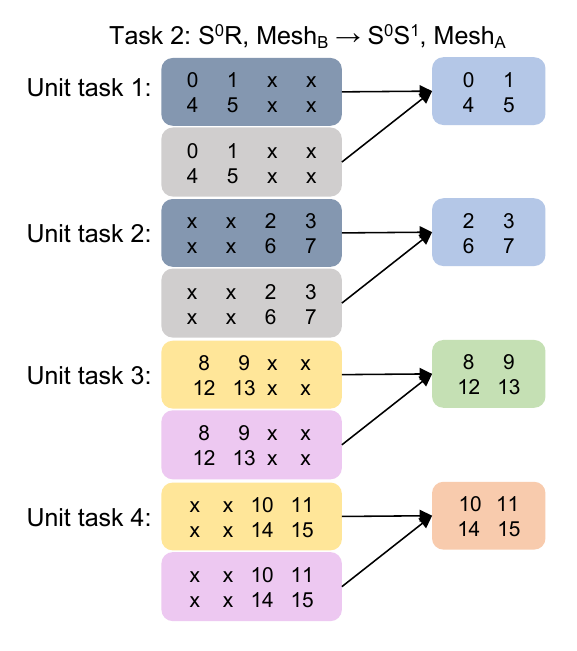}
    \vskip -1.5em
	\caption{Unit tasks for cross-mesh resharding case 2.}
    \vskip -1.5em
	\label{fig:unit_task_2}
\end{figure}
We examples of all unit tasks decomposed from cross-mesh resharding tasks in Figure~\ref{fig:device-view}. The unit tasks for cross-mesh resharding task 1 are listed in Figure~\ref{fig:unit_task_1}, and that for cross-mesh resharding tasks 2 are in Figure~\ref{fig:unit_task_2}.
\subsection{Algorithm to decompose a cross-mesh resharding}
We decompose a cross-mesh resharding into unit tasks by two steps: first, we split a tensor into slices based on the sharding spec of both sender and receiver mesh.

After that, for each slice, we create a unit task whose senders are all send devices owning the slice, and receivers are all receive devices requiring the slice.

To create slices, we enumerate the tensor dimensions. Each dimension $d$ has a set of cutpoints, denoted by $Cut(d)$. We use $l_d$ to represent the length of dimension $d$. $Cut(d)$ is initialized with value $\{0, l_d\}$ for each dimension $d$

If dimension $d$ is sharded on the sender mesh with a degree of $k$, the dimension is split into $k$ equal size replicas, and thus we add $l_d/k, 2l_d/k, \dots$ into the cutpoint set $Cut_d$. The same applies for the sharding spec on receiver dimension as well. Eventually, for a tensor with $n$ dimensions $d_0,d_1\dots d_{n-1}$, we get $n$ corresponding cutpoint sets.

Let $c_d$ be the size of the cutpoint set $Cut_d$. For each cutpoint set $Cut_d = \{p_0, p_1, \dots p_{c_d - 1}\}$ where $p_0<p_1<\dots <p_{c_d-1}$, we create the corresponding interval set $I_d = \{(p_0,p_1), (p_1,p_2)\dots\ (p_{c_d-2},p_{c_d-1})\}$. The final slice set is represented by the cross product of the interval set for all tensor dimensions, i.e. $I_0\times I_1\times\dots I_{d-1}$.

\end{document}